\documentclass{article}

\usepackage{times}
\usepackage{graphicx} 
\usepackage{subfigure}

\usepackage{natbib}

\usepackage{algorithm}
\usepackage{algorithmic}

\usepackage{amsmath}
\usepackage{amsfonts}
\usepackage{amssymb}

\newcommand{\eqn}[1]{Eq.~\ref{#1}}
\newcommand{\sect}[1]{Section~\ref{#1}}
\newcommand{\tbl}[1]{Table~\ref{#1}}
\newcommand{\fig}[1]{Figure~\ref{#1}}

\newcommand{\logb}{\log_2}
\newcommand{\nlayers}{{n_\text{layers}}}

\usepackage[accepted]{icml2014}

\begin{document}

\twocolumn[
\icmltitle{Deep AutoRegressive Networks}

\icmlauthor{Karol Gregor}{karolg@google.com}
\icmlauthor{Ivo Danihelka}{danihelka@google.com}
\icmlauthor{Andriy Mnih}{amnih@google.com}
\icmlauthor{Charles Blundell}{cblundell@google.com}
\icmlauthor{Daan Wierstra}{wierstra@google.com}
\icmladdress{Google DeepMind}

\icmlkeywords{generative models, deep learning, minimum description length}

\vskip 0.3in
]

\begin{abstract}
We introduce a deep, generative autoencoder capable of learning hierarchies of distributed representations from data.
Successive deep stochastic hidden layers are equipped with autoregressive connections, which enable the model to be sampled from quickly and exactly via ancestral sampling.
We derive an efficient approximate parameter estimation method based on the minimum
description length (MDL) principle,
which can be seen as maximising a variational lower bound on the log-likelihood, with a feedforward neural network implementing approximate inference. 
We demonstrate state-of-the-art generative performance on a number of classic data sets: several UCI data sets, MNIST and Atari 2600 games.
\end{abstract}

\section{Introduction}

Directed generative models provide a fully probabilistic account of observed random variables and their latent representations.
Typically either the mapping from observation to representation or representation to observation is intractable and hard to approximate efficiently.
In contrast, autoencoders provide an efficient two-way mapping where an encoder
maps observations to representations and a decoder maps representations back to
observations.
Recently several authors \citep{ranzato2007sparse,vincent2008extracting,vincent2011connection,rifai2012generative,bengio2013generalized}
have developed probabilistic versions of regularised autoencoders, along
with means of generating samples from such models.
These sampling procedures are often iterative, producing
correlated \emph{approximate} samples from previous approximate samples,
and as such explore the full distribution slowly, if at all.

In this paper, we introduce Deep AutoRegressive Networks (DARNs), deep
generative autoencoders that in contrast to the aforementioned models
efficiently generate independent, \emph{exact} samples via ancestral sampling.
To produce a sample, we simply perform a top-down pass through the decoding
part of our model, starting at the deepest hidden layer and sampling one unit
at a time, layer-wise.
Training a DARN proceeds by minimising the total information stored for
reconstruction of the original input, and as such follows the minimum
description length principle \citep[MDL; ][]{rissanen1978modeling}.
This amounts to backpropagating an MDL cost through the entire joint
encoder/decoder.

There is a long history of viewing autoencoders through the lens of MDL
\citep{hinton1993keeping, hinton1994autoencoders}, yet this has not previously been considered in the context
of deep autoencoders.
MDL provides a sound theoretical basis for DARN's regularisation,
whilst the justification of regularised autoencoders was not immediately
obvious.
Learning to encode and decode observations according to a compression metric
yields representations that can be both concise and irredundant from
an information theoretic point of view.
Due to the equivalence of compression and prediction, compressed
representations are good for making predictions and hence also good for generating
samples.
Minimising the description length of our model coincides exactly with
minimising the Helmholtz variational free energy, where our encoder plays the
role of the variational distribution.
Unlike many other variational learning algorithms, our algorithm
is not an expectation maximisation algorithm, but rather a stochastic
gradient descent method, jointly optimising all parameters of the autoencoder
simultaneously.

DARN and its learning algorithm easily stack, allowing ever deeper
representations to be learnt, whilst at the same time compressing the
training data --- DARN allows for alternating layers of stochastic hidden units and
deterministic non-linearities.
Each stochastic layer within DARN is autoregressive: each unit receives input
both from the preceding layer and the preceding units within the same layer.
Autoregressive structure captures much of the dependence among units within the same layer, at very
little computational cost during both learning and generation.
This is in marked contrast to other mechanisms for lateral connections, such as
introducing within-layer undirected edges, which often come at a prohibitively
high computational cost at training and/or generation time.

Recently, several authors have exploited autoregression for distribution
modelling \citep{ larochelle2011neural, gregor2011learning, uria2013deep}.
Unlike these models, DARN can have \emph{stochastic} hidden units, and
places autoregressive connections among these hidden units.
Depending upon the architecture of the network, this can yield gains in both
statistical and computational efficiency.

The remainder of the paper is structured as follows.
In \sect{sec:model} we describe the architecture of our model,
\sect{sec:compress} reviews the minimum description length principle
and its application to autoencoders.
\sect{sec:learn} describes the approximate parameter estimation
algorithm.
\sect{sec:results} has the results of our model on several data sets,
and we conclude with a brief summary in \sect{sec:conclusion}.

\section{Model Architecture}
\label{sec:model}

\begin{figure}
\centering
\includegraphics[width=\linewidth]{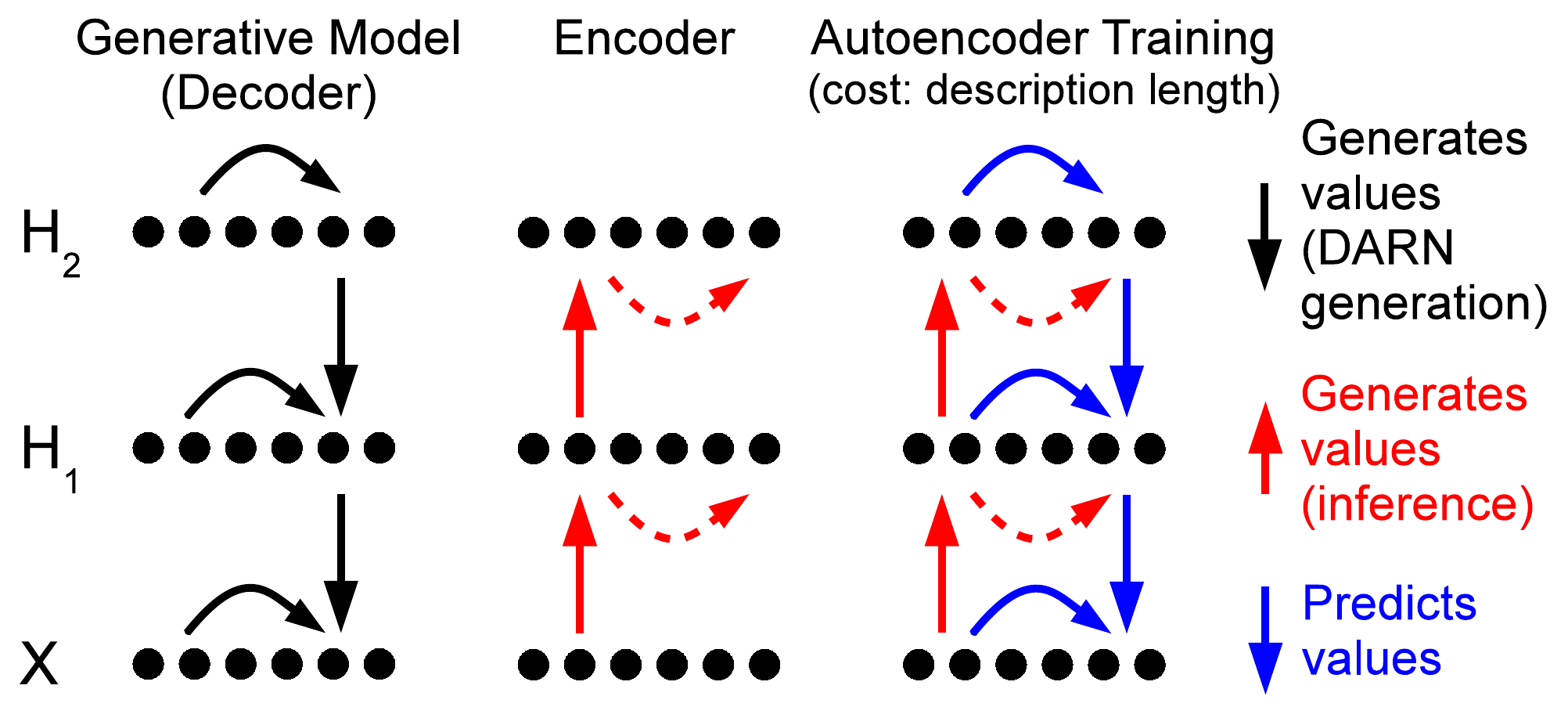}
\caption{\label{fig:autoencoder}
\textbf{Left:} DARN's decoder as a generative model.  
Top-down, ancestral sampling through DARN's decoder starts with the deepest
stochastic hidden layer $H_2$, sampling each unit in turn before proceeding
downwards to lower layers, ending by producing an observation $X$.
\textbf{Centre:} DARN's encoder as inference.
Conditioned upon the observation $X$, and sampling left-to-right, bottom-up,
DARN's encoder infers the representation $H_1, H_2$ of an observation.
\textbf{Right:} DARN as an autoencoder.
During training, the encoder infers a suitable representation $H_1, H_2$
and the decoder predicts this representation and the observation.
The parameters of the encoder and decoder are then simultaneously minimised
with respect to the implied coding cost.
This cost equals the Helmholtz variational free energy.}
\end{figure}
Our model is a deep, generative autoencoder; an example is shown in \fig{fig:autoencoder} with two hidden layers.
DARN has three components: the encoder $q(H|X)$ that picks a representation $H$
for a given observation $X$, a decoder prior $p(H)$ which provides a prior
distribution on representations $H$ for generation, and a decoder conditional
$p(X|H)$ which, given a representation $H$, produces an observation $X$.
We shall use uppercase letters for random variables and lowercase for
their values.
We shall first describe our model with a single stochastic hidden layer and
later show how this is easily generalised to multiple layers.

We begin by describing the decoder prior on the representation $h$.
The decoder prior on the representation $h$ is an autoregressive model.
Let $h_{1:j}$ denote the vector $(h_1, h_2, \dots, h_j)$
where each $h_i \in \{0,1\}$, then
\begin{align}
\label{eq:decodeprior}
p(h) &=
    \prod_{j=1}^{n_h}
    p(h_j | h_{1:j-1})
\end{align}
where $h = (h_1, h_2, \dots, h_{n_h})$ denotes the representation with
$n_h$ hidden stochastic units.
$p(h_j | h_{1:j-1})$ is the probability mass function
of $h_j$ conditioned upon the activities of the previous
units in the representation $h_{1:j-1}$.

In DARN, we parameterise the
conditional probability mass of $h_j$ in a variety of ways, depending upon
the complexity of the problem.
Logistic regression is the simplest:
\begin{align}
\label{eq:decodepriorlogreg}
p(H_j = 1 | h_{1:j-1})
&=\sigma(W^{(H)}_j \cdot h_{1:j-1} + b^{(H)}_j)
,
\end{align}
where $\sigma(x) = \frac{1}{1+e^{-x}}$.
The parameters are $W^{(H)}_j \in \mathbb{R}^{j-1}$,
which is the weight vector,
and $b^{(H)}_j \in \mathbb{R}$, which is a bias parameter.

The conditional distributions of the decoder $p(X|H)$ and of the encoder
$q(H|X)$ have similar forms:
\begin{align}
\label{eq:decodecond}
p(x|h) &=
    \prod_{j=1}^{n_x}
    p(x_j | x_{1:j-1}, h)
, \\
\label{eq:encodecond}
q(h|x) &=
    \prod_{j=1}^{n_x}
    q(h_j | h_{1:j-1}, x)
,
\end{align}
where, as with the decoder prior, the conditional probability mass functions can
be parameterised as in \eqn{eq:decodepriorlogreg} (we shall explore some
more elaborate parameterisations later).
Consequently,
\begin{align}
\label{eq:decodelogreg}
p(X_j = 1 | x_{1:j-1}, h)
&=\sigma(W^{(X|H)}_j \cdot (x_{1:j-1}, h) + b^{(X|H)}_j)
\\
\label{eq:encodelogreg}
q(H_j = 1 | h_{1:j-1}, x)
&=\sigma(W^{(H|X)}_j \cdot x + b^{(H|X)}_j)
\end{align}
where $(x_{1:j-1},h)$ denotes the concatenation of the vector $x_{1:j-1}$ with
the vector $h$, 
$W^{(X|H)}_j \in \mathbb{R}^{j-1+n_h}$
and
$W^{(H|X)}_j \in \mathbb{R}^{n_x}$
are weight vector parameters
and
$b^{(H|X)}_j$ and $b^{(X|H)}_j$ are the scalar biases.
Whilst in principle, \eqn{eq:encodelogreg} could be made autoregressive,
we shall typically choose not to do so, as this can have significant computational
advantages, as we shall see later in \sect{sec:sample}.

\subsection{Deeper Architectures}

The simple model presented so far is already a universal distribution approximator --- it can approximate any (reasonable) distribution given sufficient capacity.
As adding extra hidden layers to models such as deep belief networks strictly
improves their representational power \cite{le2008representational}, we could
ask whether that is also the case for DARN.
Although every distribution on $H$ may be written as \eqn{eq:decodeprior},
not every factorisation can be parameterised as \eqn{eq:decodepriorlogreg}.
Thus we propose boosting DARN's representational power in three ways:
by adding stochastic hidden layers,
by adding deterministic hidden layers,
and by using alternate kinds of autoregressivity.
We now consider each approach in turn.
\paragraph{Additional stochastic hidden layers.}
We consider an autoencoder with hidden stochastic layers $H^{(1)}, \ldots, H^{(\nlayers)}$ each with $n_h^{(1)},\ldots,n_h^{(\nlayers)}$ units, respectively.
For convenience we denote the input layer by $H^{(0)}=X$ and let $H^{(\nlayers+1)}=\emptyset$.
The decoder and encoder probability distributions become
\begin{align}
\label{eq:deepp}
p(H^{(l)}|H^{(l+1)})
    &=
    \prod_{j=1}^{n_h^{(l)}}
        p(H^{(l)}_j|H^{(l)}_{1:j-1}, H^{(l+1)})
,
\\
\label{eq:deepq}
q(H^{(k)}|H^{(k-1)})
    &=
    \prod_{j=1}^{n_h^{(k)}}
        q(H^{(k)}_j|H^{(k)}_{1:j-1}, H^{(k-1)})
\end{align}
for $l=0,\ldots,\nlayers$ and $k=1,\ldots,\nlayers$.

\paragraph{Additional deterministic hidden layers.}
The second way of adding complexity is to insert more complicated deterministic functions between the stochastic layers.
This applies both to the encoder and the decoder.
If we wished to add just one deterministic hidden layer,
we could use a simple multi-layer perceptron such as:
\begin{align}
d^{(l)} &= \tanh(U h^{(l+1)}) \\
p(H_j^{(l)} = 1 &| h_{1:j-1}^{(l)}, h^{(l+1)})
\nonumber \\
&=\sigma(W^{(H)}_j \cdot (h_{1:j-1}^{(l)}, d^{(l)}) + b^{(H)}_j)
.
\end{align}
where $U \in \mathbb{R}^{n_d \times n_h^{(l+1)}}$ is a weight matrix,
$n_d$ is the number of deterministic hidden units,
$W^{(H)}_j \in \mathbb{R}^{j-1+n_d}$ is a weight vector,
and $b^{(H)}_j$ is a scalar bias.

\paragraph{Alternate kinds of autoregressivity.}
Finally, we can increase representational power by using more flexible
autoregressive models, such as NADE \citep{larochelle2011neural} and
EoNADE \citep{uria2013deep}, instead of the simple linear autoregressivity
we proposed in the previous section.

The amount of information that can be stored in the representation $H$
is upper bounded by the number of \emph{stochastic} hidden units.
Additional \emph{deterministic} hidden units \emph{do not} introduce any
extra random variables and so cannot increase the capacity of the
representation,
whilst additional \emph{stochastic} hidden units \emph{can}.

\subsection{Local connectivity}
\label{sec:lcn}

DARN can be made to
scale to high-dimensional (image) data
by restricting
connectivity, both between adjacent layers, and autoregressively, within layers.
This is particularly useful for modelling larger images.
Local connectivity can be either fully convolutional
\cite{lecun1998gradient} or use less weight sharing.
In this paper we use the periodic local connectivity of
\citet{gregor2010emergence} which uses less weight sharing than full
convolutional networks.

\subsection{Sampling}
\label{sec:sample}

Sampling in DARN is simple and efficient as it is just ancestral sampling.
We start with the top-most layer, sample the first hidden unit $h_1 \sim p(H_1^{(\nlayers)})$ and then for each $i$ in turn, we sample $h_i \sim p(H_i^{(\nlayers)} | h_{1:i-1}^{(\nlayers)})$.
We repeat this procedure for each successive layer until we reach the observation.
Sampling from the encoder works in exactly the same way but in the opposite direction, sampling each hidden unit from $q(H_i^{(l)} | h_{1:i-1}^{(l)}, h^{(l-1)})$ successively.

A DARN without a stochastic hidden layer but with an autoregressive visible
layer is a fully visible sigmoid belief network \citep[FVSBN;
][]{frey1998graphical}.
Thus FVSBN sampling scales as $O(n_x^2)$.
In NADE \citep{larochelle2011neural, gregor2011learning},
autoregression is present in the visibles, but only deterministic hidden
units are used.
Sampling then scales as $O(n_x n_d)$ where $n_x$ is the number of
visibles and $n_d$ is the number of deterministic hidden units.
The complexity of sampling from a fully autoregressive single stochastic
hidden layer DARN is $O((n_h + n_x)^2)$.
If we omit the autoregressivity on the observations, we obtain
a time complexity of $O(n_h(n_x + n_h))$.
Furthermore, if the stochastic hidden layer is sparse, such that
at $n_s$ units are active on average, we obtain an expected time complexity of
$O(n_s (n_x + n_h))$.
We call this sparse, fast version fDARN.
As more stochastic or deterministic hidden layers are added to DARN, the
advantage of DARN becomes greater as each part of the decoder need only be
computed once for DARN per datum, wheras deeper NADE-like models
\citep{uria2013deep} require re-computation of large parts of the model for
each unit.

\section{Minimum Description Length and Autoencoders}
\label{sec:compress}

Autoencoders have previously been trained by an MDL principle derived
from bits-back coding \citep{hinton1993keeping}.
This yields generative autoencoders trained to minimise the Helmholtz
variational free energy \citep{hinton1994autoencoders}.
Here we extend this work to deeper, autoregressive models trained by
a stochastic approximation to backpropagation as opposed to expectation
maximisation.

According to the MDL principle, we shall train this autoencoder by finding
parameters that try to maximally compress the training data.
Suppose a sender wishes to communicate a binary sequence of $n_x$ elements,
$x$, to a receiver.
We shall first sample a representation of $n_h$ binary elements, $h$, to
communicate and then send the residual of $x$ relative to this
representation.
The idea is that the representation has a more concise code than the original
datum $x$ and so can be compressed effectively: for example,
by arithmetic coding \citep{mackay2003information}.

The description length of a random variable taking a particular value
indicates how many bits must be used to communicate that particular value.
Shannon's source coding theorem shows that
the description length is equal to the
negative logarithm of the probability of the random variable taking
that particular value \cite{mackay2003information}.
Hence, when communicating a datum $x$, having already communicated its
representation $h$, the description length would be
\begin{align}
\label{eq:decodedl}
L(x|h) &= -\logb p(x|h)
.
\end{align}
We wish for the parameters of the autoencoder to compress the data well on
average, and so we shall minimise the expected description length,
\begin{align}
\label{eq:expdl}
L(x) &= \sum_h q(h|x) (L(h) + L(x|h))
,
\end{align}
where $L(h)$ denotes the description length of the representation $h$,
and $q(h|x)$ is the encoder probability of the representation $h$.
As we are using bits-back coding,
the description length of the representation $h$ is
\begin{align}
\label{eq:priordl}
L(h) &= -\logb p(h) + \logb q(h|x)
.
\end{align}
Substituting \eqn{eq:decodedl} and \eqn{eq:priordl} into \eqn{eq:expdl} we 
recover the Helmholtz variational free energy:
\begin{align}
\label{eq:costdl}
L(x)
    &= -\sum_h q(h|x) (\logb p(x, h) - \logb q(h|x))
.
\end{align}
Picking the parameters of $q(H|X)$ and $p(X,H)$ to minimise the description
length in \eqn{eq:costdl} yields a coding scheme that requires the fewest expected
number of bits to communicate a datum $x$ and its representation $h$.

As \eqn{eq:costdl} is the variational free energy, the encoder
$q(H|X)$ that
minimises \eqn{eq:costdl} is the posterior $p(H|X)$.
Variational learning methods sometimes refer to the negative expected description length
$-L(x)$ as the expected lower bound as it serves as a lower bound upon $\logb
p(x)$.
Note here that we shall be interested in optimising the parameters of
$q(H|X)$ and $p(X,H)$ simultaneously, whereas variational learning often only
optimises the parameters of $q(H|X)$ and $p(X,H)$ by co-ordinate descent.

\section{Learning}
\label{sec:learn}

Learning in DARN amounts to jointly training weights and biases $\theta$
of both the encoder and the decoder, simultaneously, to minimise
\eqn{eq:expdl}.
The procedure is based on gradient descent by backpropagation and is based upon
a number of approximations to the gradient of \eqn{eq:expdl}.

We write the expected description length in \eqn{eq:expdl} as:
\begin{align}
\label{eq:fullloss}
L(x)
    &=
    \sum_{h_1=0}^1
    q(h_1 | x)
    \dots
    \sum_{h_{n_h}=0}^1
    q(h_{n_h}| h_{1:n_h-1}, x)
\nonumber \\&\phantom{=-\sum}
    \times (\logb q(h|x) - \logb p(x,h))
\end{align}
Calculating \eqn{eq:fullloss} exactly is intractable.
Hence we shall use a Monte carlo approximation.

Learning proceeds as follows:
\begin{enumerate}
\item Given an observation $x$, sample a representation $h \sim q(H|x)$ (see \sect{sec:sample}).
\item
Calculate $q(h|x)$ (\eqn{eq:encodecond}), $p(x|h)$ (\eqn{eq:decodecond}) and $p(h)$ (\eqn{eq:decodeprior}) for the sampled representation $h$ and
given observation $x$.
\item
Calculate the gradient of \eqn{eq:fullloss}.
\item
Update the parameters of the autoencoder by following the gradient
$\nabla_\theta L(x)$.
\item Repeat.
\end{enumerate}
We now turn to calculating the gradient of \eqn{eq:fullloss}
which requires backpropagation of the MDL cost through the joint encoder/decoder.
Unfortunately, this pass through the model includes stochastic units.
Backpropagating gradients through stochastic binary units na\"ively
yields gradients that are highly biased, yet
often work well in practice \citep{hinton2012neural}.
Whilst it is possible to derive estimators that are unbiased
\citep{bengio2013estimating}, their empirical performance is often
unsatisfactory.
In this work, we backpropagate gradients through stochastic binary units, and then
re-weight these gradients to reduce bias and variance.
Details are given in Appendix~\ref{sec:derivationOfGradients}.

Finally note that when the encoder is not autoregressive, the entire system can be trained using standard matrix operations and point-wise nonlinearities.
Hence it is easily implementable on graphical processing units.
The decoder's autoregressive computation is expressed as a full matrix multiplication with a triangular matrix.

\section{Results}
\label{sec:results}
We trained our models on binary UCI data sets, MNIST digits and frames from five Atari
2600 games \citep{bellemare12arcade}.

The quantitative results reported here are in terms of the probability the
decoder assigns to a test datum: $p(x)$.
For small DARN models, we can evaluate the likelihood $p(x)$ exactly by
iterating over every possible representation: $p(x) = \sum_h p(x,h)$.
As the computational cost of this sum grows exponentially in the size of
the representation, for DARN models with more than 16 stochastic hidden units,
we use an importance sampling estimate using the encoder distribution:
\begin{align}
\label{eq:importanceSample}
p(x) &\approx \frac{1}{S}
\sum_{s=1}^{S}
    \frac{p(x,h^{(s)})}{q(h^{(s)}|x)},
 &
h^{(s)} \sim q(H|x),
\end{align}
where $s$ indexes one of $S$ samples.
As this estimate can have high variance, we repeat the estimation ten times and
report the 95 per cent confidence interval.
In our experiments, the variance of the estimator was low.
Where available, we used a validation set to choose the learning rate
and certain aspects of the model architecture, such as the number of hidden units.
We used the Monte Carlo approximation to expected description length in
\eqn{eq:fullloss} of the validation set to select these.

\subsection{Binary UCI data sets}
We evaluated the test-set performance of DARN on eight binary data sets from the UCI repository \cite{Bache+Lichman:2013}.
In Table \ref{tab:small}, we compare DARN to baseline models from
\citet{uria2013deep}.

We used a DARN with two hidden layers.
The first layer was deterministic, with $\text{tanh}$ activations.
The second layer was a stochastic layer with an autoregressive prior $p(H)$.
The decoder conditional $p(X|H)$ included autoregressive connections.

The architecture and learning rate was selected by cross-validation on a validation set for each data set.
The number of deterministic hidden units was selected from 100 to 500, in steps
of 100,
whilst the number of stochastic hidden units was selected from $\{8,12,16,32,64,128,256\}$.
We used RMSprop \cite{graves2013generating} with momentum 0.9 and learning rates 0.00025, 0.0000675 or $10^{-5}$.
The network was trained with minibatches of size 100.
The best results are shown in bold in Table \ref{tab:small}.
DARN achieved better test log-likelihood on four of eight data sets than
the baseline models reported in \citet{uria2013deep}.
We found that regularisation by adaptive weight noise on these small data sets
\citep{graves2011practical} did not yield good results, but early stopping
based on the performance on the validation set worked well.

\begin{table*}[t]
\caption{Log likelihood (in nats) per test-set example on the eight UCI data sets.}
\label{tab:small}
\begin{center}
\begin{tabular}{l|cccccccc}
\hline
Model & Adult & Connect4 & DNA & Mushrooms & NIPS-0-12 & Ocr-letters & RCV1 & Web \\
\hline
MoBernoullis & $20.44$ & $23.41$ & $98.19$ & $14.46$ & $290.02$ & $40.56$ & $47.59$ & $30.16$ \\
RBM & $16.26$ & $22.66$ & $96.74$ & $15.15$ & $277.37$ & $43.05$ & $48.88$ & $29.38$ \\
FVSBN & $\textbf{13.17}$ & $12.39$ & $83.64$ & $10.27$ & $276.88$ & $39.30$ & $49.84$ & $29.35$ \\
NADE (fixed order) & $13.19$ & $\textbf{11.99}$ & $84.81$ & $9.81$ & $273.08$ & $\textbf{27.22}$ & $46.66$ & $28.39$ \\
EoNADE 1hl (16 ord.) & $13.19$ & $12.58$ & $\textbf{82.31}$ & $\textbf{9.68}$ & $\textbf{272.38}$ & $27.31$ & $\textbf{46.12}$ & $\textbf{27.87}$ \\
\hline
DARN & $13.19$ & $\textbf{11.91}$ & $\textbf{81.04}$ & $\textbf{9.55}$ & $274.68$ & $28.17 \pm 0$ & $\textbf{46.10} \pm \textbf{0}$ & $28.83 \pm 0$ \\
\end{tabular}
\end{center}
\vskip -0.1in
\end{table*}

\subsection{Binarised MNIST data set}
We evaluated the sampling and test-set performance of DARN on the
binarised MNIST data set \cite{salakhutdinov2008quantitative}, which consists
of $50,000$ training, $10,000$ validation, and $10,000$ testing images of
hand-written digits \cite{larochelle2011neural}.
Each image is $28 \times 28$ pixels.

\begin{figure*}[t]
\begin{center}

\vspace{.4cm}

\begin{minipage}{0.66\textwidth}
\includegraphics[width=.99\textwidth]{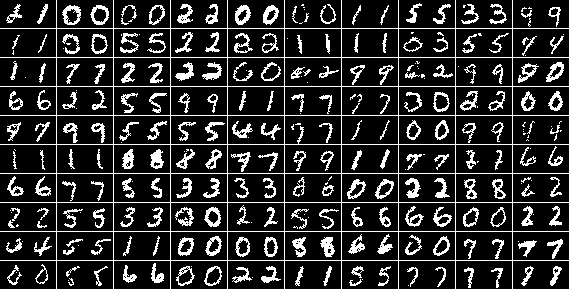}
\end{minipage}
\begin{minipage}{0.33\textwidth}
\includegraphics[width=.99\textwidth]{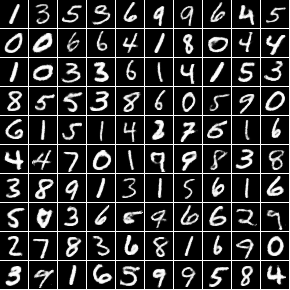}
\end{minipage}

\end{center}
\caption{\textbf{Left:} Samples from DARN paired with the nearest training example from binarised MNIST. The generated samples are not simple memorisation of the training examples.
\textbf{Right:} Sample probabilities from DARN trained on pixel intensities directly.
}
\label{fig:gendigits}
\end{figure*}

We used two hidden layers, one deterministic, one stochastic.
The
results are in \tbl{tab:binmnist} with $n_h$ denoting the number of stochastic
hidden units.
The deterministic layer had 100 units for architectures with 16 or fewer stochastic units per layer, and 500 units for more than 16 stochastic units.
The deterministic activation function was taken to be the $\text{tanh}$ function.
We used no autoregressivity for the observation layer --- the
decoder conditional is a product of independent Bernoulli distributions,
conditioned upon the representation.
Training was done with RMSprop \citep{graves2013generating}, momentum 0.9 and
minibatches of size 100.
We used a learning rate of 3$\times10^{-5}$.
Adaptive weight noise \citep{graves2011practical}, denoted by ``adaNoise'' in
\tbl{tab:binmnist}, was used to avoid the need for early stopping.

After training, we were able to measure the exact log-likelihood for networks with 16 or fewer stochastic hidden units.
For the network with 500 hidden units, we estimated the log-likelihood by importance sampling given by the above procedure.
For each test example, we sampled $100,000$ latent representations from the encoder distribution.
The estimate was repeated ten times; we report the estimated 95 per cent confidence intervals.
The obtained log-likelihoods and confidence intervals are given in \tbl{tab:binmnist} along with those of other models.
DARN performs favourably compared to the other models.
For example, a DARN with just 9 stochastic hidden units obtains almost the same
log-likelihood as a mixture of Bernoullis (MoBernoullis) with 500 components:
$\log_2 500 \approx 9$.
DARN with 500 stochastic hidden units compares favourably to
state-of-the-art generative performance of 
deep Boltzmann machines \citep[DBM; ][]{salakhutdinov2009deep}
and deep belief networks \citep[DBN; ][]{salakhutdinov2008quantitative,murray2009evaluating}.
Notably, DARN's upper bound, the expected description length given in the far
right column, is lower than the likelihood of NADE whilst DARN's estimated log
likelihood is lower than the log-likelihood of the best reported EoNADE
\citep{uria2013deep} results.

We performed several additional experiments.
First, we trained fDARN with 400 hidden units and $5\%$ sparsity, resulting in an upper bound negative log-likelihood of 96.1 (\tbl{tab:binmnist}).
The estimated speed of generation was $2.4 \times 10^4$ multiplications per sample, which compares favourably to NADE's of $2.2 \times 10^6$, a nearly 100 fold speedup.
While the likelihood is worse, the samples appear reasonable by ocular
inspection.

Next, we trained a very deep, $12$ stochastic layer DARN with $80$ stochastic units in each layer, and $400$ tanh units in each deterministic layer.
Here we also used skip connections where each tanh layer received input from all previous stochastic layers.
Due to computational constraints we only evaluated the upper bound of this architecture
as reported in \tbl{tab:binmnist} --- where it records the best upper bound among all DARN models, showing the value of depth in DARN.

Finally, we trained a network with one stochastic layer, 400 units, and one tanh layer (1000 units) in both encoder and decoder on the pixel intensities directly,
rather than binarising the data set.
We show the sample probabilities of observables in \fig{fig:gendigits}(right).

\begin{table}[t]
\caption{Log likelihood (in nats) per test-set example on the binarised MNIST
data set.
The right hand column, where present, represents the expected description length
(in \eqn{eq:fullloss}) which serves as an upper bound on the log-likelihood
for DARN.}
\label{tab:binmnist}
\begin{center}
\begin{tabular}{l|cccccccc}
\hline
Model & $-\log p$ & $\le$ \\
\hline
MoBernoullis K=10 & $168.95$ & \\
MoBernoullis K=500 & $137.64$ & \\
RBM (500 h, 25 CD steps) & $\approx 86.34$ & \\
DBM 2hl & $\approx 84.62$ &\\
DBN 2hl & $\approx \textbf{84.55}$ &\\
NADE 1hl (fixed order) & $88.33$ & \\
EoNADE 1hl (128 orderings) & $87.71$ & \\
EoNADE 2hl (128 orderings) & $85.10$ & \\
\hline
DARN $n_h$=9, adaNoise & $138.84$ & $145.36$ \\
DARN $n_h$=10, adaNoise & $133.70$ & $140.86$ \\
DARN $n_h$=16, adaNoise & $122.80$ & $130.94$ \\
DARN $n_h$=500 & $84.71 \pm 0.01$ & $90.31$ \\
DARN $n_h$=500, adaNoise & $\textbf{84.13} \pm \textbf{0.01}$ & $88.30$ \\
\hline
fDARN $n_h=400$ & - & $96.1$ \\
deep DARN & - & \textbf{87.72} \\
\end{tabular}
\end{center}
\vskip -0.1in
\end{table}

\subsection{Atari 2600 game frames}

\begin{figure*}[t]
\begin{minipage}{.6\linewidth}
\centering
\includegraphics[width=\linewidth]{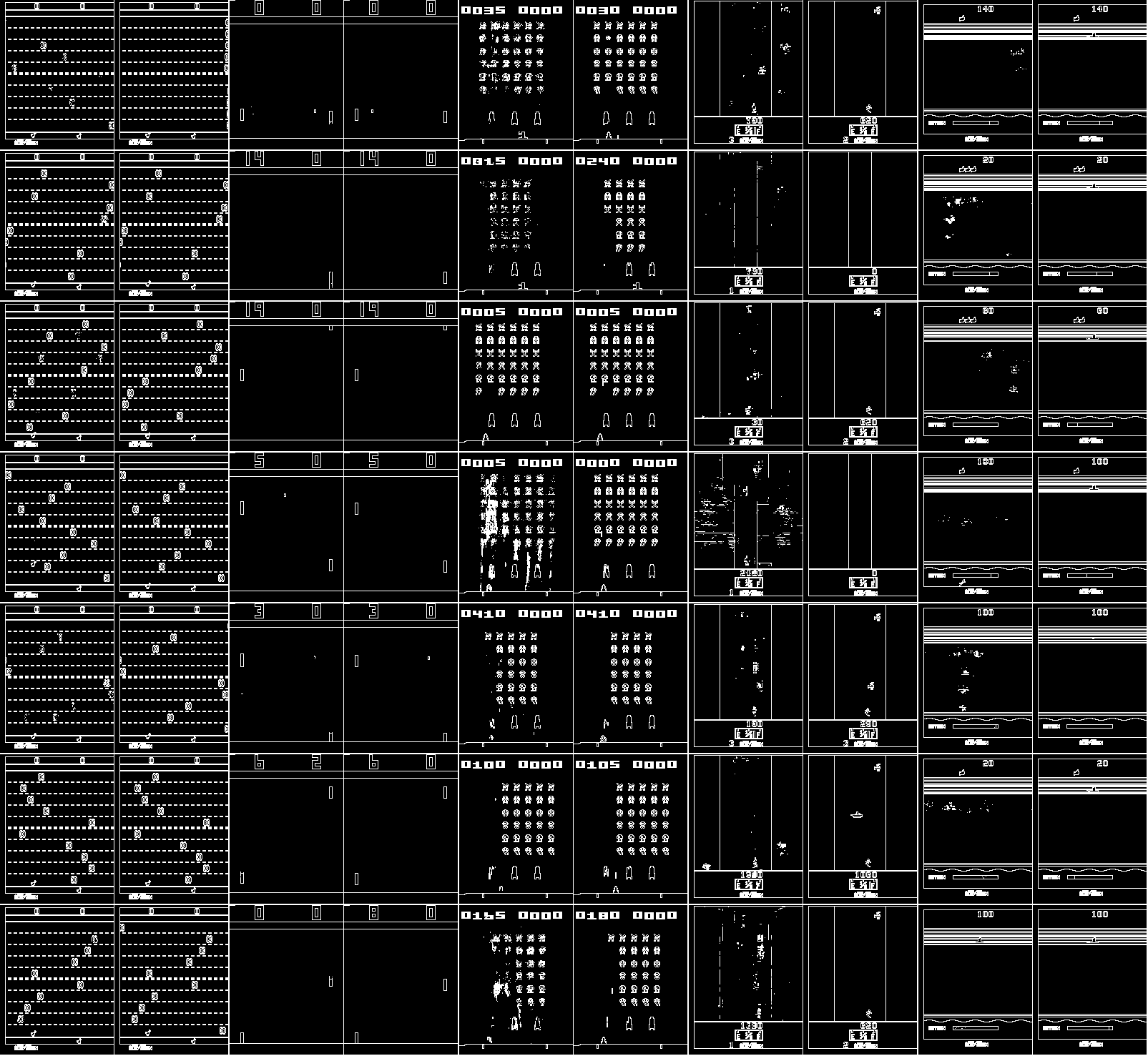}
\caption{\label{fig:atari}Samples from a locally connected DARN paired with the nearest training example. Some of the generated frames have novel combinations of objects and scores.
Columns from left to right, along with upper bound negative log-likelihood on the test set: Freeway (19.9), Pong (23.7), Space Invaders (113.0), River Raid (139.4), Sea Quest (217.9).
}
\end{minipage}
\hspace{1ex}
\centering
\begin{minipage}{.35\linewidth}
\begin{center}
\begin{minipage}{0.9\textwidth}
\begin{center}
\includegraphics[width=.2\textwidth]{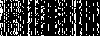}
\end{center}
\end{minipage}
\begin{minipage}{0.9\textwidth}
\begin{center}
\includegraphics[width=.3\textwidth]{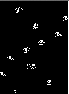}
\end{center}
\end{minipage}
\begin{minipage}{0.9\textwidth}
\begin{center}
\includegraphics[width=.4\textwidth]{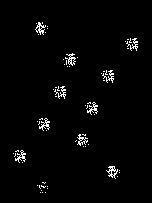}
\end{center}
\end{minipage}
\begin{minipage}{0.9\textwidth}
\begin{center}
\includegraphics[width=.5\textwidth]{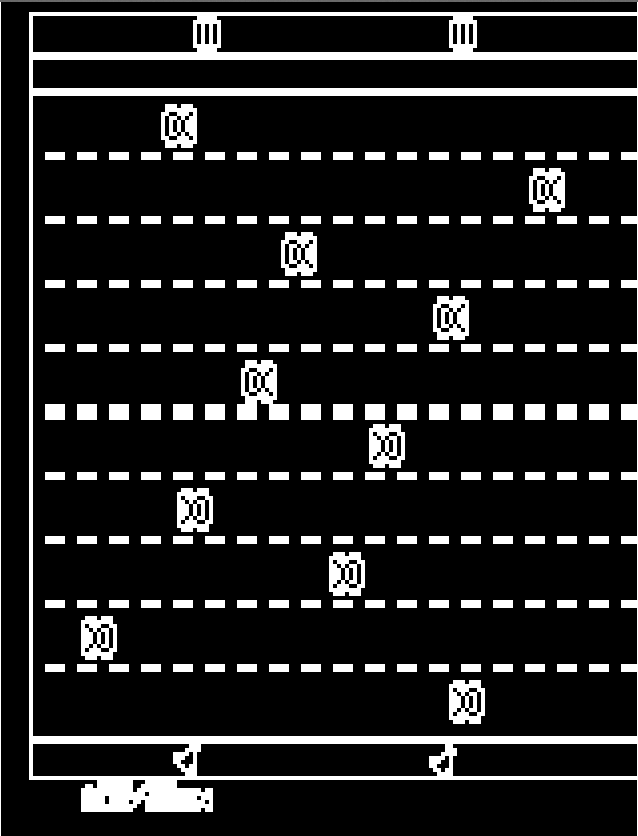}
\end{center}
\end{minipage}
\end{center}\caption{\label{fig:reprfreeway}
\textbf{Bottom:} An input frame from the game Freeway.
\textbf{Lower Middle:}
Activations in the encoder from the first hidden layer.
\textbf{Upper Middle:}
Activations in the encoder from the second hidden layer.
\textbf{Top:}
Each of $25$ rows is an activation of the fully-connected 
third hidden layer with $100$ units.
}
\end{minipage}
\end{figure*}

We recorded $100,000$ frames from five different Atari 2600 games \citep{bellemare12arcade} from random
play, recording each frame with $1\%$ probability.
We applied an edge detector to the images yielding frames of $159\times 209$
pixels\footnote{The script to generate the dataset is available at \texttt{https://github.com/fidlej/aledataset}}.
Frames of these games generated by DARN are shown in \fig{fig:atari}.

\begin{figure}[t]
\begin{center}
\begin{minipage}{0.8\textwidth}
\includegraphics[width=.2\textwidth]{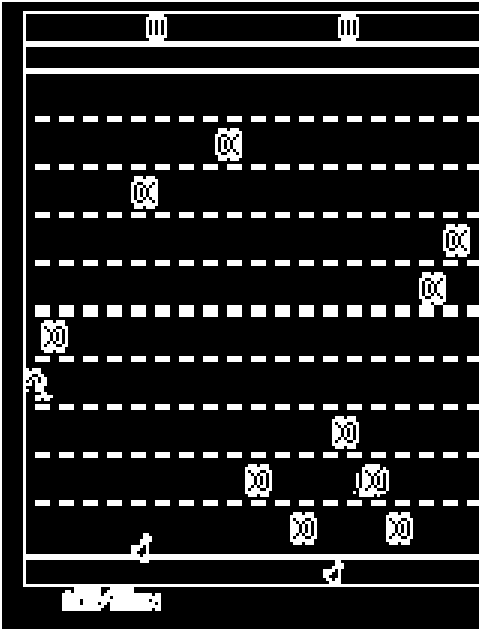}
\includegraphics[width=.2\textwidth]{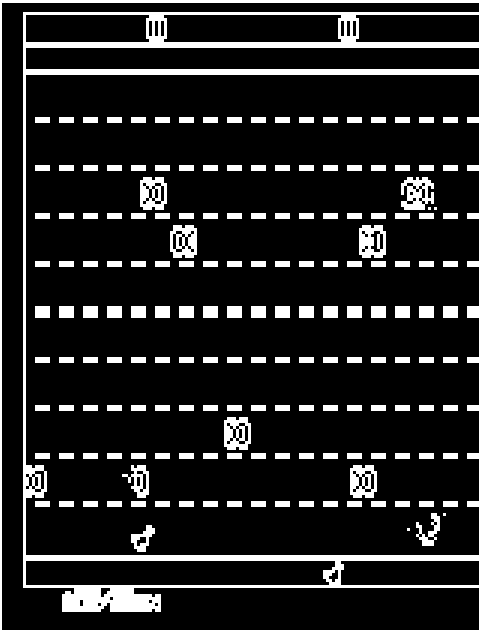}
\includegraphics[width=.2\textwidth]{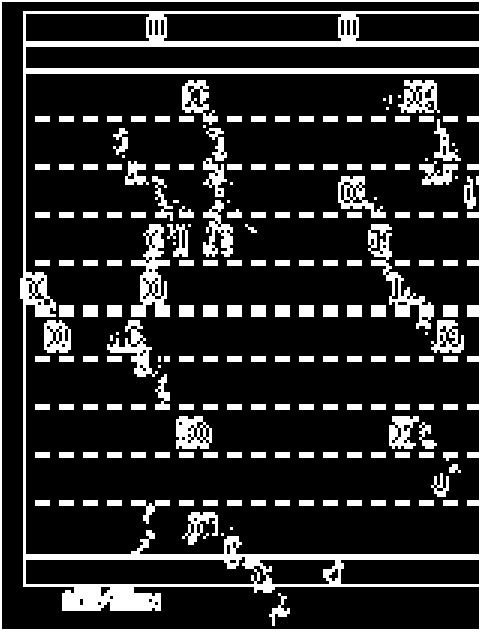}
\end{minipage}
\end{center}
\caption{\label{fig:genfreeway}
Samples from networks of different depths. 
\textbf{Left:}
Frames generated from the network described in \fig{fig:reprfreeway}.
\textbf{Middle:}
Frames generated from the network with the same sizes but without the fully-connected top layer.
\textbf{Right:}
Frames generated from the network with the top two layers removed.  }
\end{figure}

To scale DARN to these larger images, we used three hidden layers.
The stack of layers contains a locally connected layer with weight-sharing (see \sect{sec:lcn}), a rectified linear activation function,
another locally connected layer followed by a rectified linear function
and a fully connected layer followed by 300 stochastic binary units.
The autoregressive prior on the representation was also fully connected.
The locally connected layers had 32 filters with a period of 8.
The first locally connected layer used stride 4 and kernel size 8.
The second locally connected layer used stride 2 and kernel size 4.
The autoregressive connections for the visible layer used a period of 14
and kernel size 15.

The games in \fig{fig:atari} are ordered from left to right, in decreasing
order of log probability.
While DARN captures much of the structure in these games, such as
the scores, various objects, and correlations between objects, when
the game becomes less regular, as with River Raid (second to right) and Sea Quest (far right), DARN is not able to generate reasonable frames.

\fig{fig:reprfreeway} shows the representation that DARN learns 
for a typical frame of the game Freeway. 
To show the effect of deeper layers in DARN, \fig{fig:genfreeway} shows the
frames DARN generates using the representation from different depths.
In the game of Freeway, several cars (the white blobs) travel along lanes, delimited
by dashed white lines.
Here we used a DARN with three stochastic hidden layers.
When DARN was trained using a sparsity penalty on the activations, as we described
in \sect{sec:sample}, it learnt a representation where the second hidden layer
captures the rough outline of each car, whereas the first hidden layer filled
in the details of each car.
A global bias learns the background image.
A third hidden layer decides where to place the cars.
All layers except the very top layer are locally connected.
The top layer has 100 units.
The second hidden layer was locally connected with a kernel of size $21\times 21$,
whilst the first hidden layer was locally connected with a kernel of size $7\times 7$.

\section{Conclusion}
\label{sec:conclusion}

In this paper we introduced deep autoregressive networks (DARN), a new deep generative architecture with autoregressive stochastic hidden units capable of capturing high-level structure in data to generate high-quality samples. The method, like the ubiquitous autoencoder framework, is comprised of not just a decoder (the generative element) but a stochastic encoder as well to allow for efficient and tractable inference.
Training proceeds by backpropagating an MDL cost through the joint model, which approximately equates to minimising the Helmholtz variational free energy.
This procedure necessitates backpropagation through stochastic units, as such yielding an approximate Monte Carlo method.
The model samples efficiently, trains efficiently and is scalable to locally connected and/or convolutional architectures.
Results include state-of-the-art performance on multiple data sets.

\appendix

\section{Derivation of Gradients}
\label{sec:derivationOfGradients}

We derive the gradient of the objective function
with respect to the inputs to a stochastic \textit{binary} hidden unit.
Let $q(h_i)$ be the probability distribution of the $i$th hidden unit,
$f(h_i)$ be the part of the network which takes $h_i$ as an input.
The expected value of $f(h_i)$ and its gradient are:
\begin{align}
\mathbb{E}\left[f(H_i)\right] &= \sum_{h_i=0}^1 q(h_{i}) f(h_i)
\\
    \nabla_\theta \mathbb{E}\left[f(H_i)\right] &=
\sum_{h_i=0}^1
    q(h_i)
\nabla_\theta \log q(h_i) f(h_i)
\end{align}
where $f$ does not depend directly on the inputs $\theta$ of the $q(h_i)$ distribution.
We use a Monte Carlo approximation to estimate
$\nabla_\theta \mathbb{E}\left[f(H_i)\right]$ where $h_{i}$ is sampled from $q(h_i)$
\citep{williams1992simple,andradottir1998review}.
Monte Carlo approximations of the gradient are unbiased but can have high variance.
To combat the high variance, we introduce a baseline $b$, inspired by control variates \cite{paisley2012variational}:
\begin{align}
\label{eq:withBaseline}
\nabla_\theta \mathbb{E}\left[f(H_i)\right]
& \approx \mathbb{E}\left[\widehat{\nabla_\theta F}\right]
\\
\label{eq:estimator}
\widehat{\nabla_\theta F} & =
    \nabla_\theta \log q(h_i) (f(h_i) - b)
\end{align}
where $\widehat{\nabla_\theta F}$ denotes our estimator.
A good baseline should be correlated with $f(h_i)$, have low variance, and also
be such that the expected value of $\nabla_\theta \log q(h_i) b$ is zero to get
an unbiased estimate of the gradient.
We chose a \textit{non-constant} baseline.
The baseline will be a first-order Taylor approximation of $f$.
We can get the first-order derivatives from backpropagation.
The baseline is a Taylor approximation of $f$
about $h_i$, evaluated at a point $h_i^\prime$:

\begin{align}
b(h_i) &= f(h_i) + \frac{df(h_i)}{dh_i} (h_i^\prime - h_i)
\end{align}

To satisfy the unbiasedness requirement, we need to solve
the following equation for $h_i^\prime$:

\begin{align}
0 &= \sum_{h_i=0}^1
    q(h_i)
    \nabla_\theta \log q(h_i) (
    f(h_i) + \frac{df(h_i)}{dh_i} (h_i^\prime - h_i))
\end{align}

The solution depends on the shape of $f$.
If $f$ is a linear function, any $h_i^\prime$ can be used.
If $f$ is a quadratic function, the solution is $h_i^\prime = \frac{1}{2}$.
If $f_i$ is a cubic or higher-order function, the solution depends on the coefficients of the polynomial.
We will use $h_i^\prime = \frac{1}{2}$ and our estimator will be biased for non-quadratic functions.

By substituting the baseline into \eqn{eq:estimator} we obtain the final form of our estimator of the gradient:
\begin{align}
\widehat{\nabla_\theta F} &=
    \nabla_\theta \log q(h_i) \frac{df(h_i)}{dh_i} (h_i - \frac{1}{2})
\\
    &= \frac{\nabla_\theta q(H_i=1)}{2q(h_i)}\frac{df(h_i)}{dh_i}
\end{align}

An implementation can estimate
the gradient with respect to $q(H_i=1)$
by backpropagating with respect to $h_i$ and scaling the gradient by $\frac{1}{2q(h_i)}$, where $h_i$ is the sampled binary value.
\bibliographystyle{icml2014}
\bibliography{darn_arxiv}

\begin{thebibliography}{28}
\providecommand{\natexlab}[1]{#1}
\providecommand{\url}[1]{\texttt{#1}}
\expandafter\ifx\csname urlstyle\endcsname\relax
  \providecommand{\doi}[1]{doi: #1}\else
  \providecommand{\doi}{doi: \begingroup \urlstyle{rm}\Url}\fi

\bibitem[Andrad{\'o}ttir(1998)]{andradottir1998review}
Andrad{\'o}ttir, Sigr{\'u}n.
\newblock A review of simulation optimization techniques.
\newblock In \emph{Simulation Conference Proceedings, 1998. Winter}, volume~1,
  pp.\  151--158. IEEE, 1998.

\bibitem[Bache \& Lichman(2013)Bache and Lichman]{Bache+Lichman:2013}
Bache, K. and Lichman, M.
\newblock {UCI} machine learning repository, 2013.
\newblock URL \url{http://archive.ics.uci.edu/ml}.

\bibitem[{Bellemare} et~al.(2013){Bellemare}, {Naddaf}, {Veness}, and
  {Bowling}]{bellemare12arcade}
{Bellemare}, M.~G., {Naddaf}, Y., {Veness}, J., and {Bowling}, M.
\newblock The arcade learning environment: An evaluation platform for general
  agents.
\newblock \emph{Journal of Artificial Intelligence Research}, 47:\penalty0
  253--279, 06 2013.

\bibitem[Bengio et~al.(2013{\natexlab{a}})Bengio, L{\'e}onard, and
  Courville]{bengio2013estimating}
Bengio, Yoshua, L{\'e}onard, Nicholas, and Courville, Aaron.
\newblock Estimating or propagating gradients through stochastic neurons for
  conditional computation.
\newblock \emph{arXiv preprint arXiv:1308.3432}, 2013{\natexlab{a}}.

\bibitem[Bengio et~al.(2013{\natexlab{b}})Bengio, Yao, Alain, and
  Vincent]{bengio2013generalized}
Bengio, Yoshua, Yao, Li, Alain, Guillaume, and Vincent, Pascal.
\newblock Generalized denoising auto-encoders as generative models.
\newblock \emph{arXiv preprint arXiv:1305.6663}, 2013{\natexlab{b}}.

\bibitem[Frey(1998)]{frey1998graphical}
Frey, Brendam~J.
\newblock \emph{Graphical models for machine learning and digital
  communication}.
\newblock The MIT press, 1998.

\bibitem[Graves(2011)]{graves2011practical}
Graves, Alex.
\newblock Practical variational inference for neural networks.
\newblock In \emph{Advances in Neural Information Processing Systems}, pp.\
  2348--2356, 2011.

\bibitem[Graves(2013)]{graves2013generating}
Graves, Alex.
\newblock Generating sequences with recurrent neural networks.
\newblock \emph{arXiv preprint arXiv:1308.0850}, 2013.

\bibitem[Gregor \& LeCun(2010)Gregor and LeCun]{gregor2010emergence}
Gregor, Karol and LeCun, Yann.
\newblock Emergence of complex-like cells in a temporal product network with
  local receptive fields.
\newblock \emph{arXiv preprint arXiv:1006.0448}, 2010.

\bibitem[Gregor \& LeCun(2011)Gregor and LeCun]{gregor2011learning}
Gregor, Karol and LeCun, Yann.
\newblock Learning representations by maximizing compression.
\newblock \emph{arXiv preprint arXiv:1108.1169}, 2011.

\bibitem[Hinton(2012)]{hinton2012neural}
Hinton, Geoffrey.
\newblock Neural networks for machine learning.
\newblock Coursera video lectures, 2012.

\bibitem[Hinton \& Van~Camp(1993)Hinton and Van~Camp]{hinton1993keeping}
Hinton, Geoffrey~E and Van~Camp, Drew.
\newblock Keeping the neural networks simple by minimizing the description
  length of the weights.
\newblock In \emph{Proceedings of the sixth annual conference on Computational
  learning theory}, pp.\  5--13, 1993.

\bibitem[Hinton \& Zemel(1994)Hinton and Zemel]{hinton1994autoencoders}
Hinton, Geoffrey~E and Zemel, Richard~S.
\newblock Autoencoders, minimum description length, and helmholtz free energy.
\newblock \emph{Advances in Neural Information Processsing Systems}, pp.\
  3--3, 1994.

\bibitem[Larochelle \& Murray(2011)Larochelle and Murray]{larochelle2011neural}
Larochelle, Hugo and Murray, Iain.
\newblock The neural autoregressive distribution estimator.
\newblock \emph{Journal of Machine Learning Research}, 15:\penalty0 29--37,
  2011.

\bibitem[Le~Roux \& Bengio(2008)Le~Roux and Bengio]{le2008representational}
Le~Roux, Nicolas and Bengio, Yoshua.
\newblock Representational power of restricted boltzmann machines and deep
  belief networks.
\newblock \emph{Neural computation}, 20\penalty0 (6):\penalty0 1631--1649,
  2008.

\bibitem[LeCun et~al.(1998)LeCun, Bottou, Bengio, and
  Haffner]{lecun1998gradient}
LeCun, Yann, Bottou, L{\'e}on, Bengio, Yoshua, and Haffner, Patrick.
\newblock Gradient-based learning applied to document recognition.
\newblock \emph{Proceedings of the IEEE}, 86\penalty0 (11):\penalty0
  2278--2324, 1998.

\bibitem[MacKay(2003)]{mackay2003information}
MacKay, David~JC.
\newblock \emph{Information theory, inference and learning algorithms, Ch 6}.
\newblock Cambridge university press, 2003.

\bibitem[Murray \& Salakhutdinov(2009)Murray and
  Salakhutdinov]{murray2009evaluating}
Murray, Iain and Salakhutdinov, Ruslan.
\newblock Evaluating probabilities under high-dimensional latent variable
  models.
\newblock In \emph{Advances in Neural Information Processing Systems}, pp.\
  1137--1144. 2009.

\bibitem[Paisley et~al.(2012)Paisley, Blei, and Jordan]{paisley2012variational}
Paisley, John, Blei, David, and Jordan, Michael.
\newblock Variational bayesian inference with stochastic search.
\newblock \emph{arXiv preprint arXiv:1206.6430}, 2012.

\bibitem[Ranzato et~al.(2007)Ranzato, Boureau, and Cun]{ranzato2007sparse}
Ranzato, Marc'Aurelio, Boureau, Y-lan, and Cun, Yann~L.
\newblock Sparse feature learning for deep belief networks.
\newblock In \emph{Advances in Neural Information Processing Systems}, pp.\
  1185--1192, 2007.

\bibitem[Rifai et~al.(2012)Rifai, Bengio, Dauphin, and
  Vincent]{rifai2012generative}
Rifai, Salah, Bengio, Yoshua, Dauphin, Yann~N, and Vincent, Pascal.
\newblock A generative process for sampling contractive auto-encoders.
\newblock In \emph{Proceedings of the 29th International Conference on Machine
  Learning (ICML-12)}, pp.\  1855--1862, 2012.

\bibitem[Rissanen(1978)]{rissanen1978modeling}
Rissanen, Jorma.
\newblock Modeling by shortest data description.
\newblock \emph{Automatica}, 14\penalty0 (5):\penalty0 465--471, 1978.

\bibitem[Salakhutdinov \& Hinton(2009)Salakhutdinov and
  Hinton]{salakhutdinov2009deep}
Salakhutdinov, Ruslan and Hinton, Geoffrey~E.
\newblock Deep boltzmann machines.
\newblock In \emph{International Conference on Artificial Intelligence and
  Statistics}, pp.\  448--455, 2009.

\bibitem[Salakhutdinov \& Murray(2008)Salakhutdinov and
  Murray]{salakhutdinov2008quantitative}
Salakhutdinov, Ruslan and Murray, Iain.
\newblock On the quantitative analysis of deep belief networks.
\newblock In \emph{Proceedings of the 25th International Conference on Machine
  Learning}, pp.\  872--879. ACM, 2008.

\bibitem[Uria et~al.(2013)Uria, Murray, and Larochelle]{uria2013deep}
Uria, Benigno, Murray, Iain, and Larochelle, Hugo.
\newblock A deep and tractable density estimator.
\newblock \emph{arXiv preprint arXiv:1310.1757}, 2013.

\bibitem[Vincent(2011)]{vincent2011connection}
Vincent, Pascal.
\newblock A connection between score matching and denoising autoencoders.
\newblock \emph{Neural computation}, 23\penalty0 (7):\penalty0 1661--1674,
  2011.

\bibitem[Vincent et~al.(2008)Vincent, Larochelle, Bengio, and
  Manzagol]{vincent2008extracting}
Vincent, Pascal, Larochelle, Hugo, Bengio, Yoshua, and Manzagol,
  Pierre-Antoine.
\newblock Extracting and composing robust features with denoising autoencoders.
\newblock In \emph{Proceedings of the 25th International Conference on Machine
  Learning}, pp.\  1096--1103, 2008.

\bibitem[Williams(1992)]{williams1992simple}
Williams, Ronald~J.
\newblock Simple statistical gradient-following algorithms for connectionist
  reinforcement learning.
\newblock \emph{Machine learning}, 8\penalty0 (3-4):\penalty0 229--256, 1992.

\end{thebibliography}

\end{document}